\documentclass[10pt,conference]{IEEEtran}

\IEEEoverridecommandlockouts

\usepackage{times}
\usepackage{graphicx}
\usepackage{booktabs}
\usepackage{amsmath}
\usepackage{amssymb}
\usepackage{xcolor}
\usepackage{url}
\usepackage[noadjust]{cite}

\usepackage[capitalize]{cleveref}
\usepackage{microtype}
\usepackage{tikz}
\usetikzlibrary{shapes.geometric,arrows.meta,positioning,fit,calc,backgrounds,decorations.pathreplacing}
\definecolor{Blue}{HTML}{2196F3}
\definecolor{Green}{HTML}{4CAF50}
\definecolor{Red}{HTML}{D32F2F}
\definecolor{Gray}{HTML}{666666}

\title{CAX-Agent: A Lightweight Agent Harness for Reliable APDL Automation}
\author{
\IEEEauthorblockN{Chenying Lin$^{1,6,*}$, Yichen Hai$^{1,2,*}$, Yi He$^{1}$, Ran Wang$^{2}$, Haiyan Qiang$^{2}$, Liang Yu$^{3,4,5}$}
\IEEEauthorblockA{$^{1}$Shanghai Ultradimension Technology Co., Ltd., Shanghai, China}
\IEEEauthorblockA{$^{2}$College of Logistics Engineering, Shanghai Maritime University, Shanghai, 201306, China}
\IEEEauthorblockA{$^{3}$School of Civil Aviation, Northwestern Polytechnical University, Xi'an 710072, China}
\IEEEauthorblockA{$^{4}$State Key Laboratory of Airliner Integration Technology and Flight Simulation, Shanghai 200126, China}
\IEEEauthorblockA{$^{5}$National Key Laboratory of Strength and Structural Integrity, Xi'an 710065, China}
\IEEEauthorblockA{$^{6}$Wuhan University, Wuhan, Hubei 430072, China}
\IEEEauthorblockA{$^{*}$Co-first authors with equal contribution}
}

\begin{document}
\maketitle

\begin{abstract}
Large language models deployed for MAPDL finite-element simulation face practical reliability challenges: without structured execution control, tool encapsulation, and fault recovery, outputs may be inconsistent and task failures are common. The Agent Harness paradigm addresses this by inserting domain-specific orchestration middleware that manages tool lifecycles, workflow state, and recovery escalation. This paper presents the architecture of CAX-Agent, a lightweight agent harness purpose-built for MAPDL automation, and empirically evaluates one of its core components---the recovery policy. CAX-Agent organizes execution into three layers---LLM service, agent harness, and solver backend---with a recovery ladder that escalates from deterministic rule patching through model-driven regeneration to context enrichment and human intervention. We evaluate three recovery strategies (no\_recovery, rule\_only, and model\_only) on 50 standard structural benchmarks with three repeated runs per strategy (450 case-runs total). Two independent human raters score task completion under blind conditions; inter-rater agreement is strong (quadratic weighted Cohen's kappa = 0.84, 96 percent of score pairs within one point). Model\_only achieves the best completion rate (0.9267), task score (3.59/4), total score (9.16/10), and zero-intervention rate (0.84), outperforming rule\_only (0.7733, 3.17/4, 7.03/10, 0.00) and no\_recovery (0.6933, 2.74/4, 5.60/10, 0.00) with large effect sizes (Cliff's delta = 0.81--0.87). The benchmark uses deliberately simple geometries to isolate recovery-policy effects; we discuss the scope of these findings and directions for broader validation.
\end{abstract}


\section{Introduction}
Computer-aided technologies are often grouped under the CAX umbrella, where CAD, CAE, and CAM represent design, engineering analysis, and manufacturing planning, respectively. In this work, the implemented pipeline enables CAD-plus-CAE automation only; CAM execution is out of scope.

LLM-driven finite-element simulation requires more than accurate code generation. The Transformer architecture established the self-attention paradigm for sequence modeling \cite{ref01}, and deep bidirectional pre-training extended this to representation learning \cite{ref02}. Scaling to 175B parameters enabled few-shot learning without task-specific fine-tuning \cite{ref03}, and interleaving reasoning traces with tool-use actions improved multi-step task completion \cite{ref04}. These advances have enabled tool-using code agents, but in engineering simulation, pre-processing, solver execution, and post-processing must chain correctly, and runtime errors---meshing failures, convergence issues, missing results---are common even for structurally simple tasks. Without explicit recovery mechanisms, a single failure terminates the pipeline. As LLM-based engineering agents move toward practical use, the question of how to design and evaluate recovery policies becomes central to system reliability. CAX-Agent is designed natively for MAPDL rather than adapting a general agent framework; its recovery logic is tightly coupled to MAPDL error log syntax and APDL script structure, following a rules-first, model-second escalation strategy where deterministic rule-based patches are attempted before invoking LLM-driven repair.

The Agent Harness paradigm has emerged as the core architectural pattern for bridging this gap. Rather than expecting the LLM to manage its own execution, a harness inserts a domain-specific orchestration middleware that integrates skill encapsulation, tool orchestration, workflow checkpoints, state management, and fault diagnosis with retry escalation. This middleware provides the engineering skeleton that the LLM alone cannot supply. Analysis of 70 agent-system projects identified five recurring design dimensions---scheduler type, planning capability, recovery mechanism, context management, and implementation complexity---with Agent Loop-based schedulers remaining dominant \cite{ref05}. KAIJU, an executive kernel that decouples tool execution from LLM reasoning with intent-gated execution, demonstrated that this separation enforces behavioral guarantees that prompting alone cannot match \cite{ref06}. In parallel, LLM-driven agents have been applied across engineering domains. CAD automation and generative design-to-manufacturing pipelines have been explored \cite{ref07,ref08}, alongside design structure generation and self-cognitive product design systems \cite{ref09,ref10}. End-to-end CFD automation with structured knowledge and reasoning has been demonstrated \cite{ref11}. Broader surveys cover next-generation CAE opportunities and the manufacturing lifecycle \cite{ref12,ref13}, as well as vision-language evaluation for engineering design and AI-empowered CAE \cite{ref14,ref15}; detailed discussion is deferred to Section II. These works advance agent capabilities in specific domains but do not evaluate the recovery component of a harness under controlled, repeated conditions with human-judged outcomes.

This paper presents CAX-Agent, a lightweight, native agent harness purpose-built for APDL automation in mechanical simulation. Rather than adapting a generic harness framework, CAX-Agent is designed around the specific failure patterns observed in MAPDL execution: meshing failures, convergence errors, element-type mismatches, and missing post-processing results. Its architecture separates LLM service, harness orchestration, and solver backend into three layers, with a recovery ladder that escalates from deterministic rule patching through model-driven script regeneration to context enrichment and human intervention as a final fallback. The orchestrator---not the LLM---owns retry budgets, tool dispatch, and stop conditions.

We evaluate three recovery strategies under an identical benchmark protocol: no\_recovery (one-shot execution), rule\_only (deterministic rule-based patching), and model\_only (LLM-driven error-log-conditioned regeneration with bounded retries). The benchmark uses 50 standard structural tasks---beams, plates, and cylinders under static, modal, and thermal loading---with three repeated runs per strategy (450 case-runs total). The tasks are deliberately simple. Our aim is not to push the complexity frontier of autonomous simulation, but to isolate the effect of recovery-policy design in a setting where the base task is well within the model's capability, so that outcome differences can be attributed to the recovery policy rather than to task difficulty. We report completion behavior, multi-axis scoring (human-assessed task quality plus system-derived autonomy and efficiency), and pairwise statistical tests. Model\_only achieves the strongest reliability while preserving high autonomy in this setting.

Our contributions are: (1)~CAX-Agent, a lightweight, MAPDL-native agent harness with a three-layer architecture and recovery ladder, designed around real MAPDL failure patterns; (2)~a controlled, repeated-run comparison of three recovery strategies on 50 standardized APDL tasks, with blind human scoring and inter-rater validation; and (3)~empirical evidence that model-driven recovery substantially outperforms rule-based repair in both completion rate and zero-intervention rate, with per-type failure analysis showing where residual errors concentrate.

\begin{figure*}[t]
  \centering
  \includegraphics[width=\textwidth]{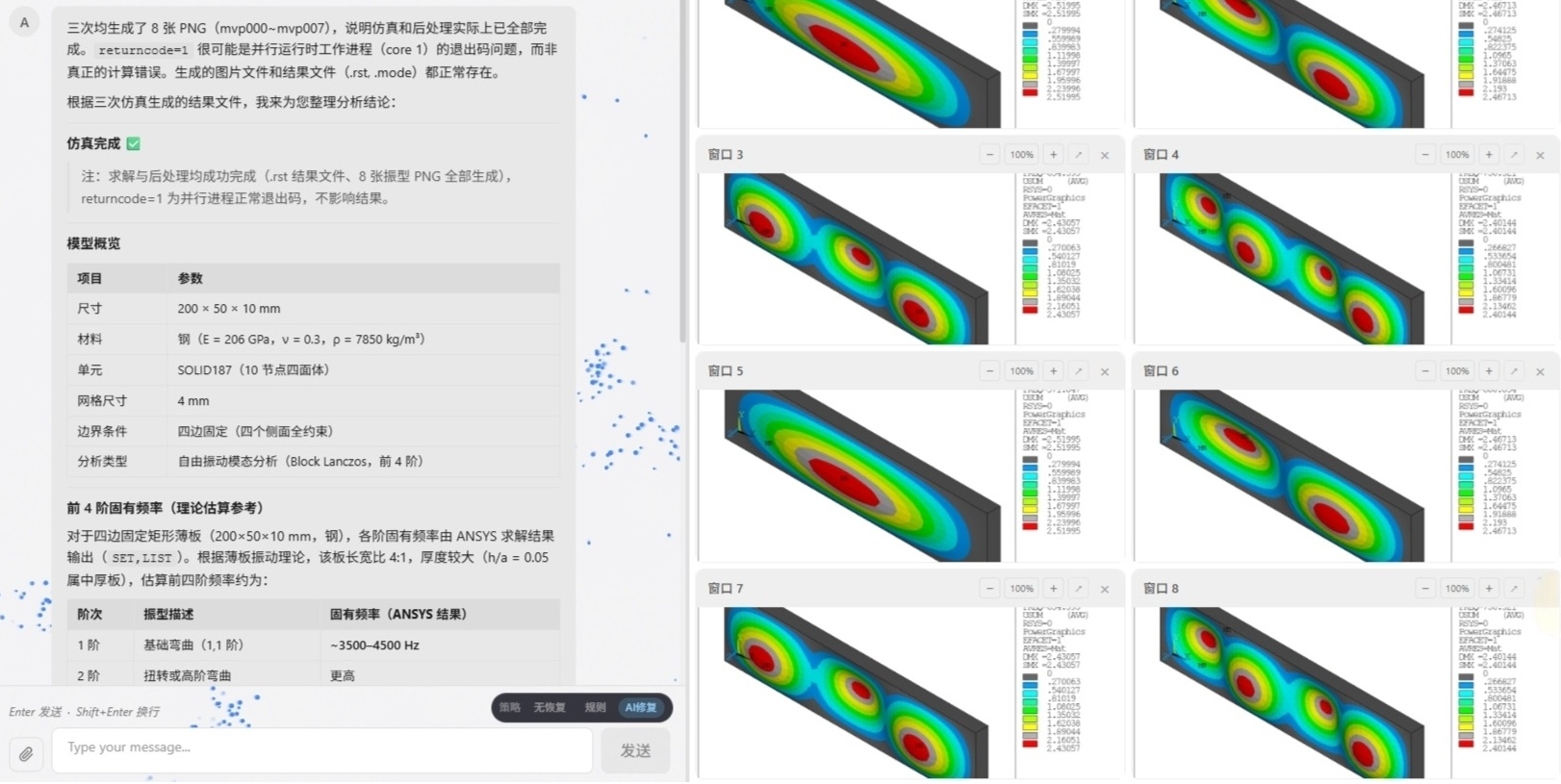}
  \caption{End-to-end UI example from a representative modal analysis run. The system autonomously generates the APDL script, executes it in MAPDL, and produces post-processing images with a conversational interface.}
  \label{fig:ui_modal_example}
\end{figure*}

Figure~\ref{fig:ui_modal_example} shows a representative end-to-end execution from the CAX-Agent interface, illustrating a conversational modal analysis task with autonomous APDL generation, MAPDL execution, and post-processing output.

\section{Related Work}
\subsection{LLM-Based Tool Use and Engineering Automation}
Tool-using LLM agents increasingly combine reasoning traces with external actions, enabling non-trivial multi-step workflows. These capabilities are now being transferred to engineering informatics. Xu et al. substantially reduced process planning construction time using an LLM-enabled knowledge graph method \cite{ref16}, and Stathatos et al. framed high-level process planning as a sequence prediction task for GPT-2 in distributed manufacturing \cite{ref17}. Shi et al. fine-tuned an LLM for automated building-code compliance \cite{ref18}. Wen et al. proposed an LLM-based human-machine collaborative approach for diagnosing complex industrial equipment faults \cite{ref19}. Zhang et al. applied a knowledge-graph-enhanced LLM to hydraulic structure safety question answering \cite{ref20}, and Wang et al. applied multimodal LLMs to construction safety inspection \cite{ref21}. These capabilities are directly relevant to simulation automation, where script generation must interact with strict solver interfaces and runtime feedback, as demonstrated in multi-agent aerodynamic optimization \cite{ref22} and surveyed for industrial embodied intelligence \cite{ref23}.

In code-oriented settings, model outputs can be strong yet brittle when execution constraints are strict. Guo et al. outlined next-generation LLM-enabled CAE opportunities \cite{ref12}, Li et al. surveyed LLMs across the manufacturing lifecycle \cite{ref13}, and Picard et al. evaluated vision-language models from conceptual design through manufacturing \cite{ref14}---all reporting that reliability under runtime constraints motivates explicit recovery controls in the agent loop.

\subsection{LLM-Driven Finite Element Automation}
Recent work has explored LLM-driven finite element automation from multiple angles. Mudur et al. proposed FEABench, benchmarking one-shot and agent-loop LLM capability on COMSOL multiphysics tasks and reporting that executable API call generation reaches 88\% but full problem completion remains challenging \cite{ref24}. Hou et al. presented AutoFEA, improving FEA input file accuracy through a GCN-Transformer retrieval model integrated with LLM planning, evaluated on CalculiX-derived benchmarks \cite{ref25}.

These studies advance generation quality and pipeline coverage under diverse conditions. Our work complements them by isolating recovery-policy design as a controlled variable: we keep the task set, model, and solver fixed while varying only the recovery strategy, with repeated-run statistics and multi-axis scoring. To our knowledge, no prior study reports such a controlled head-to-head comparison of recovery configurations for APDL automation.

\subsection{Agent Execution Infrastructure}
Beyond the engineering simulation domain, a parallel line of work addresses the infrastructure layer for LLM agents---the harness that manages tool lifecycles, retry logic, error propagation, and execution traces. Wei characterizes the dominant Agent Loop as a single-ready-unit scheduler and proposes Graph Harness, which separates planning, execution, and recovery into independent layers with a formalized node state machine \cite{ref05}. Guerin and Guerin introduce KAIJU, an executive kernel that decouples tool execution from LLM reasoning, with Intent-Gated Execution for security and configurable execution modes for different task complexities \cite{ref06}. These systems share a key design principle with CAX-Agent: the orchestrator---not the LLM---owns retry budgets, tool dispatch, and stop conditions. Where our work differs is in the empirical focus: rather than proposing a new harness architecture, we study how a specific harness component (the recovery policy) behaves under controlled conditions with repeated measurements and human evaluation.

\section{Methodology}
\subsection{System Architecture}
CAX-Agent is organized as a three-layer stack for APDL-centric execution. In CAX terms, the benchmark enables CAD-plus-CAE tasks: CAD-oriented prompt interpretation and geometry/simulation script construction, followed by CAE execution and validation through MAPDL.

\textbf{Layer 1 (routing layer).} A FastAPI-based entrypoint maintains a module registry and routes each request by module key. Incoming requests are validated against registered modules and dispatched to the corresponding sub-agent handler. This layer is responsible for function-level traffic routing across registered modules.

\textbf{Layer 2 (local lightweight model layer).} The runtime invokes a local inference backend for fast first-pass APDL generation and repair loop calls before returning tool actions to the orchestrator. In the deployed CAX setup, this layer runs Qwen-27B as the local model.

\textbf{Layer 3 (unified external LLM API layer).} External model access is unified behind a gateway configuration that manages authentication and base-URL routing. The experiment protocol fixes the external model to Claude Sonnet 4.6. This layer provides the high-capability API completion path when local reasoning is insufficient.

Above the three layers, the orchestrator converts user instructions to APDL scripts, triggers MAPDL execution, collects logs, and coordinates bounded repair attempts. A connector layer selects available solver backends (PyMAPDL, CLI MAPDL, or fallback mode) while maintaining a unified simulation interface; retry budgets and iteration traces are recorded for post-analysis.

\begin{figure*}[t]
  \centering
  \begin{tikzpicture}[
    scale=1.15,
    box/.style={rectangle,rounded corners=3pt,line width=1.2pt,fill=white,
      minimum width=2.8cm,minimum height=0.9cm,align=center,font=\small\sffamily},
    smallbox/.style={rectangle,rounded corners=2pt,line width=0.8pt,fill=white,
      minimum width=2.2cm,minimum height=0.6cm,align=center,font=\footnotesize\sffamily},
    tinybox/.style={rectangle,rounded corners=1.5pt,line width=0.6pt,fill=white,
      minimum width=1.8cm,minimum height=0.45cm,align=center,font=\scriptsize\sffamily},
    arr/.style={->,>=Stealth,line width=1.0pt,color=black!55},
    darr/.style={->,>=Stealth,line width=0.8pt,dashed},
    lbl/.style={font=\footnotesize\itshape\color{black!45}},
    section/.style={font=\bfseries\sffamily},
    every node/.style={inner sep=3pt},
  ]

  \node[section,Blue] at (1.5,8.5) {LLM Service};
  \node[section,Green] at (1.5,4.4) {Agent Harness};
  \node[section,Red] at (1.5,-0.3) {Solver Backend};

  \node[box,draw=Blue] (router) at (3.8,7.4) {Routing\\\scriptsize Module Registry (FastAPI)};
  \node[box,draw=Blue] (local) at (8.5,7.4) {Local LLM\\\scriptsize Qwen-27B, first-pass};
  \node[box,draw=Blue] (ext) at (13.2,7.4) {External LLM API\\\scriptsize Claude Sonnet 4.6};

  \draw[arr] (router) -- (local) node[lbl,below,midway,yshift=-2pt] {dispatch};
  \draw[arr] (local) -- (ext)  node[lbl,below,midway,yshift=-2pt] {escalate};

  \node[box,draw=Blue,minimum height=0.6cm,font=\footnotesize\sffamily] (prompt) at (3.8,6.0) {User Prompt};
  \draw[arr] (prompt) -- (router);

  \node[smallbox,draw=Green] (ctx)   at (3.8,5.0) {Context Manager\\\scriptsize Compress \textbar{} Trim \textbar{} Collapse};
  \node[smallbox,draw=Green] (tool)  at (3.8,3.8) {Tool Pipeline\\\scriptsize Validate $\to$ Permit $\to$ Execute};
  \node[smallbox,draw=Green] (state) at (3.8,2.6) {State Tracker\\\scriptsize Message Pairing Invariant};

  \node[box,draw=Green,minimum width=3.4cm,minimum height=2.2cm] (orch) at (8.5,3.8) {Orchestrator Core\\\scriptsize ReAct Loop (while true)\\ \scriptsize Retry Budget \textbar{} Stop Control\\\scriptsize Execution Trace \textbar{} Checkpoint};

  \node[smallbox,draw=Red] (guard) at (13.2,5.0) {Exception Guard\\\scriptsize Withhold \& Self-Repair};
  \node[smallbox,draw=Red] (reclbl) at (13.2,3.8) {Recovery Ladder};
  \node[tinybox,draw=Gray] (l1) at (13.2,3.0) {L1: Rule Patch (free)};
  \node[tinybox,draw=Gray] (l2) at (13.2,2.3) {L2: LLM Regen (cheap)};
  \node[tinybox,draw=Gray] (l3) at (13.2,1.6) {L3: Context Enrich (paid)};
  \node[tinybox,draw=Gray] (l4) at (13.2,0.9) {L4: Human Escalation};

  \draw[arr] (ext.south) -- ++(0,-0.3) -| (orch.north);
  \draw[arr] (ctx) -- (orch);
  \draw[arr] (tool) -- (orch);
  \draw[arr] (state) -- (orch);

  \draw[arr] (orch) -- (guard);
  \draw[arr] (guard) -- (reclbl);
  \draw[arr] (reclbl) -- (l1);
  \draw[arr] (l1) -- (l2);
  \draw[arr] (l2) -- (l3);
  \draw[arr] (l3) -- (l4);

  \node[box,draw=Red,minimum width=3.4cm,minimum height=1.2cm] (mapdl) at (8.5,-0.3) {MAPDL Engine\\\scriptsize PyMAPDL \textbar{} CLI \textbar{} Fallback};
  \node[smallbox,draw=Red,minimum width=2.4cm] (errlog) at (13.2,-0.3) {Error Log Extractor};

  \draw[arr] (orch) -- (mapdl);
  \draw[arr] (mapdl) -- (errlog);

  \node[box,draw=Green,minimum height=0.6cm,font=\footnotesize\sffamily] (out) at (8.5,-1.6) {Post-Processing Image Output};
  \draw[arr] (mapdl) -- (out);

  \draw[darr,Red] (errlog.east) -- ++(1.2,0) |- (reclbl.east);
  \node[font=\footnotesize\itshape\color{Red},right=4pt] at ($(reclbl.east)+(0.3,-0.2)$) {error feedback};

  \draw[darr,Gray] (l4.west) -- ++(-5.5,0) |- ($(orch.south)+(0,-0.15)$)
    node[lbl,pos=0.55,below=2pt] {bounded retries};

  \draw[darr,Gray] (orch.west) -- ++(-1.0,0) |- ($(router.north)+(0,0.3)$) -- (router.north)
    node[lbl,pos=0.7,left=2pt] {regenerate APDL};

  \end{tikzpicture}
  \caption{CAX-Agent runtime architecture showing the three-layer harness design with recovery policy selection and feedback loops.}
  \label{fig:runtime_arch}
\end{figure*}
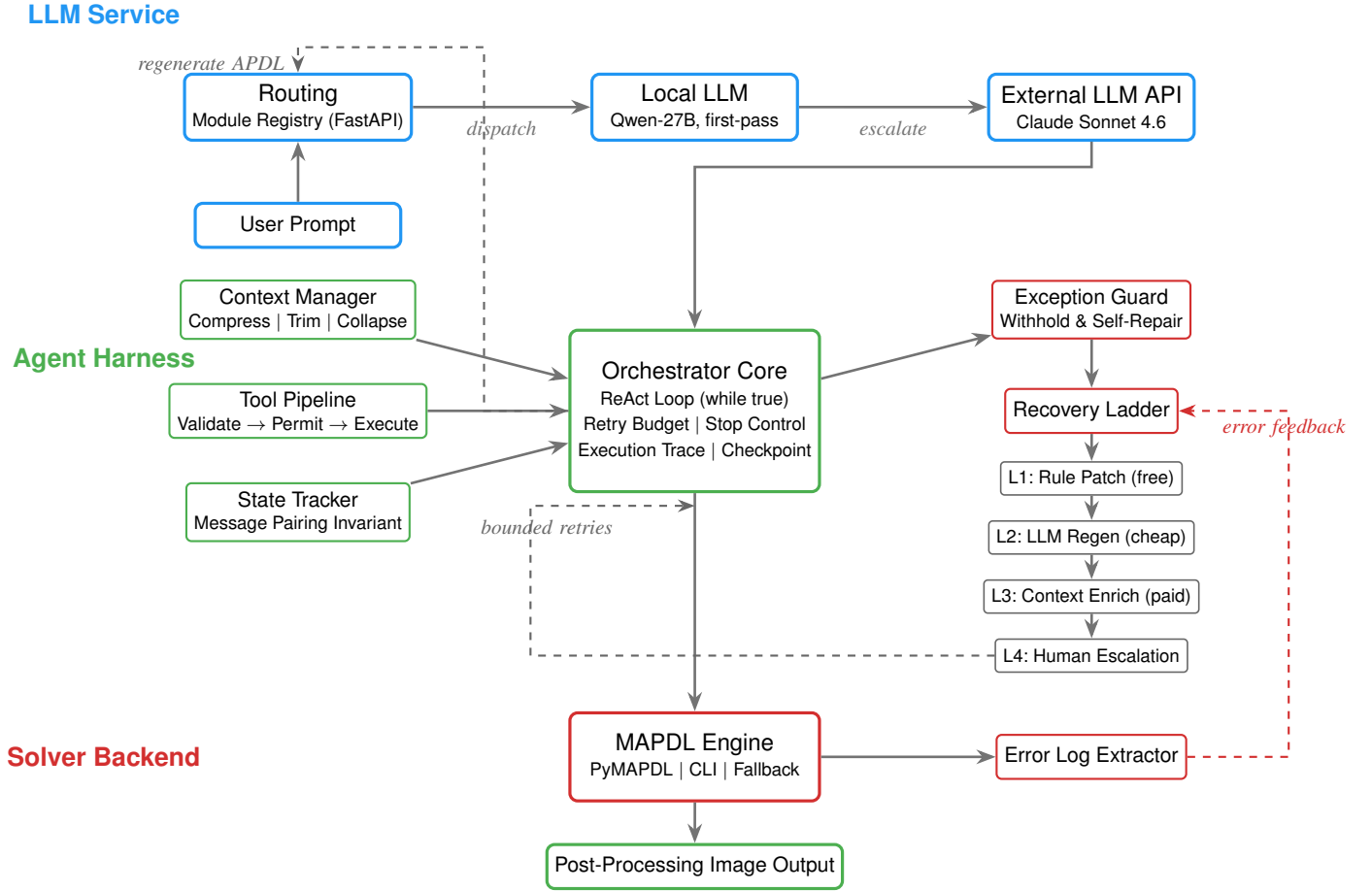

Figure~\ref{fig:runtime_arch} summarizes the loop. A failed execution emits solver logs that are re-injected into the model prompt for targeted regeneration. This design separates generation from execution control: the model handles semantic repair while the orchestrator enforces retry budgets and stop conditions.

\subsection{Implementation Details}
The LLM temperature was fixed at 0 throughout all experiments to eliminate stochastic variation. A typical successful case-run consumed approximately 10--15K tokens for the initial APDL generation; recovery attempts consumed additional tokens proportional to the number of retries but below the linear scaling factor of the retry count, because error-log-augmented prompts reused cached context prefixes. The full system runs on a single workstation, with the local Qwen-27B model serving fast first-pass generation and the external Claude Sonnet 4.6 API providing high-capability repair when the local model is insufficient or recovery is required. The rule set was derived from an internal engineering analysis of common MAPDL failure modes observed during system development.

\subsection{Recovery Strategy Definitions}
We compare three strategies under identical task sets and run counts:
\begin{itemize}
  \item \textbf{no\_recovery}: one-shot execution without repair. The agent is limited to 2 ReAct iterations (the minimum required for one generate-then-execute cycle in the tool-use framework) with no forced retries, and the error-log reading tool is disabled. If the single execution fails, the task is marked as failed.
  \item \textbf{rule\_only}: deterministic log-to-patch correction without additional model reasoning. After the initial LLM-generated APDL fails, the system reads the MAPDL error log and applies four deterministic string-transform rules: (1)~mesh failure triggers element-size doubling and free-mesh fallback; (2)~convergence failure inserts multi-substep and auto time-stepping directives before \texttt{SOLVE}; (3)~element-type errors substitute compatible element formulations; (4)~missing post-processing results rewrite \texttt{SET} commands to target the last available load step. The patched script is re-executed once. The agent performs up to 12 ReAct reasoning steps but receives no forced script retries and cannot read error logs for model-driven repair. The retry budget $B=2$ refers to APDL execution attempts (initial plus one rule-based retry), not to ReAct reasoning steps.
  \item \textbf{model\_only}: error-log-conditioned model regeneration with bounded retries. The agent has access to the error-log reading tool and runs up to 12 ReAct iterations. If the model attempts to stop after a failed simulation, the orchestrator forces up to 3 additional retry rounds, instructing the model to read the error log, diagnose the failure, and regenerate the APDL script.
\end{itemize}

A task is counted as successfully completed when the pipeline produces at least one post-processing image output from MAPDL execution.

Formally, for a given case, let $G_0$ be the initial APDL script generated by the LLM from the user prompt. If execution of $G_0$ succeeds (produces an image), the case terminates. If it fails with solver error log $e_1$, the recovery policy $\pi$ produces a revised script $G_1 = \pi(G_0, e_1)$. In general, after the $t$-th failure, $G_t = \pi(G_{t-1}, e_t)$. The process repeats until either success or a strategy-specific retry budget $B$ is exhausted:
\begin{itemize}
  \item \textbf{no\_recovery}: $B=1$, $\pi=\varnothing$ (no repair; fail on first error).
  \item \textbf{rule\_only}: $B=2$, $\pi = f_{\text{rule}}(G, e)$ where $f_{\text{rule}}$ applies up to four deterministic string-transform rules based on patterns in $e$.
  \item \textbf{model\_only}: $B=4$ (one initial attempt plus up to three forced retries), $\pi = f_{\text{LLM}}(G, e)$ where the LLM reads $e$ and regenerates the APDL script.
\end{itemize}
The budget $B$ is enforced by the orchestrator, not by the LLM. This separation---the model proposes repairs, the harness enforces budgets---is the key architectural property under test.

\subsection{Metrics and Statistical Testing}
For each case-run $i$, we construct three scores. A human rater assigns a task completion score $t_i \in \{0,1,2,3,4\}$ by inspecting the output image for correctness and completeness. The system derives an autonomy score $a_i \in \{0,1,2,3\}$ from execution logs (3 = fully autonomous, 0 = required manual intervention) and a recovery efficiency score $e_i \in \{0,1,2,3\}$ from the retry count and solver outcome. The composite total score for the run is
\begin{equation}
q_i = t_i + a_i + e_i, \qquad q_i \in [0,10].
\end{equation}
Binary completion $c_i \in \{0,1\}$ is set to 1 iff the pipeline produced at least one post-processing image. From these per-run values we compute, for each strategy $s$, the completion rate $R_s$, mean total score $Q_s$, and zero-intervention rate $Z_s = \frac{1}{N}\sum_{i} \mathbf{1}[a_i=3]$, each aggregated over $N=150$ case-runs.

For pairwise strategy comparisons we use Cliff's $\delta$, a non-parametric effect size defined for two independent samples $X = \{x_1,\dots,x_n\}$ and $Y = \{y_1,\dots,y_m\}$ as
\begin{equation}
\delta(X,Y) = \frac{1}{nm}\sum_{i=1}^{n}\sum_{j=1}^{m} \operatorname{sgn}(x_i - y_j),
\end{equation}
where $\operatorname{sgn}(z)$ is $+1$, $-1$, or $0$ for $z>0$, $z<0$, or $z=0$. Cliff's $\delta$ ranges from $-1$ to $+1$ and is directly interpretable: $\delta = P(X>Y) - P(X<Y)$. We follow the convention that $|\delta|<0.147$ is negligible, $<0.33$ small, $<0.474$ medium, and $\geq 0.474$ large. We report two-sided Mann--Whitney U $p$-values alongside $\delta$, and 95\% binomial confidence intervals on each completion rate.

The rule\_only strategy records zero-intervention rate 0.00 because its deterministic patches require a confirmation step before re-execution, which is logged as a non-autonomous action; the strategy is rules-first with human fallback rather than automated model repair, and the zero reflects this design property, not an absence of recovery activity.

We note that case-runs sharing the same task are correlated by design (three repeats per task). As a sensitivity check, we repeated all pairwise comparisons using per-task mean scores; the between-strategy ranking and effect-size directions are unchanged. The reported results treat the 150 runs as independent observations, consistent with standard practice for controlled repeated-measure benchmarks.

Two raters independently scored $t_i$ for all 450 case-runs under blind conditions. Inter-rater agreement is measured by quadratic weighted Cohen's $\kappa$, defined as
\begin{equation}
\kappa_w = \frac{p_o - p_e}{1 - p_e},
\end{equation}
where $p_o$ is the observed proportion of weighted agreement and $p_e$ is the expected proportion under chance, using squared-error weights between ordinal categories.

\section{Experiments}
\subsection{Benchmark Scope}
Our benchmark consists of 50 APDL simulation prompts spanning three categories: 35 static analysis tasks (beams, plates, brackets, pressure vessels, and simple assemblies), 10 modal analysis tasks, and 5 steady-state thermal analysis tasks. All geometries are standard structural elements with regular cross-sections and well-defined boundary conditions. The tasks are deliberately simple in their physics: they involve linear elasticity, small deformations, and single-physics setups. This design choice is intentional---by keeping the simulation domain tractable, we can attribute differences in outcomes primarily to recovery-policy design rather than to the inherent difficulty of the simulation problem. The benchmark is not intended to represent the full complexity of industrial CAE workflows; it is a testbed for comparing agent recovery behavior under controlled conditions.

\subsection{Benchmark Protocol}
Two raters independently scored task completion under blind conditions. Quadratic weighted Cohen's $\kappa = 0.84$, indicating excellent agreement; 96\% of score pairs fall within one point. Both raters produce the same strategy ranking (model\_only $>$ rule\_only $>$ no\_recovery). Rater~B's means are systematically 0.3--0.5 points lower, reflecting a stricter threshold, but between-strategy gaps remain consistent. All reported results use Rater~A scores; Rater~B statistics confirm the same conclusions. The per-case scoring data is available from the authors upon reasonable request.

\subsection{Main Results}
Table~\ref{tab:overall} reports the core metrics. Model\_only leads on all indicators: completion rate 0.9267 (95\% CI: 0.885--0.968), task score 3.59/4, total score 9.16/10, and zero-intervention 0.84. Completion rates for the baselines are substantially lower: 0.7733 (95\% CI: 0.706--0.840) for rule\_only and 0.6933 (95\% CI: 0.620--0.767) for no\_recovery; the confidence intervals for model\_only do not overlap with those of either baseline.

Pairwise effect sizes, measured by Cliff's $\delta$ on the total score distribution, are large: $\delta = 0.81$ for model\_only vs. rule\_only, $\delta = 0.87$ for model\_only vs. no\_recovery, and $\delta = 0.57$ for rule\_only vs. no\_recovery (all $p < 0.001$, Mann--Whitney U). An effect of this magnitude means that a randomly chosen model\_only case-run outscores a rule\_only case-run with probability $(0.81+1)/2 \approx 0.91$.

On the human-scored task completion axis alone, model\_only outperforms rule\_only by +0.42 points and no\_recovery by +0.85 points (Mann--Whitney U, $p=2.74\times10^{-3}$ and $p=1.85\times10^{-4}$). Rule\_only vs. no\_recovery on task score is not significant ($p=0.173$), indicating that deterministic rule patching improves system-level autonomy metrics but does not reliably raise output quality as judged by human raters.

\begin{table}[t]
\centering
\caption{Overall benchmark results (150 case-runs per strategy). Task score is human-assessed; autonomy and efficiency are system-derived and folded into the total.}
\label{tab:overall}
\begin{tabular}{lcccc}
\toprule
Strategy & Compl.\ Rate & Task (/4) & Total (/10) & Zero-Interv. \\
\midrule
model\_only  & 0.9267 & 3.5867 & 9.1600 & 0.8400 \\
rule\_only   & 0.7733 & 3.1667 & 7.0267 & 0.0000 \\
no\_recovery & 0.6933 & 2.7400 & 5.6000 & 0.0000 \\
\bottomrule
\end{tabular}
\end{table}

\begin{figure}[t]
  \centering
  \includegraphics[width=\linewidth]{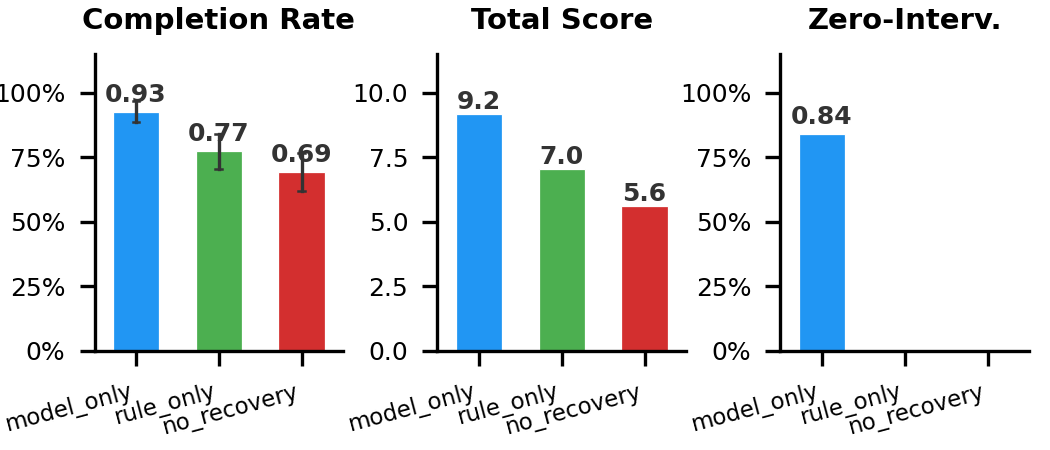}
  \caption{Comparison of completion rate, scaled average total score, and zero-intervention rate across the three recovery strategies. The model\_only strategy consistently leads all metrics.}
  \label{fig:overall_compare}
\end{figure}

Figure~\ref{fig:overall_compare} visualizes these results across the three primary metrics, highlighting the consistent lead of model\_only.

\subsection{Per-Type Analysis and Robustness}
\begin{figure}[t]
  \centering
  \includegraphics[width=\linewidth]{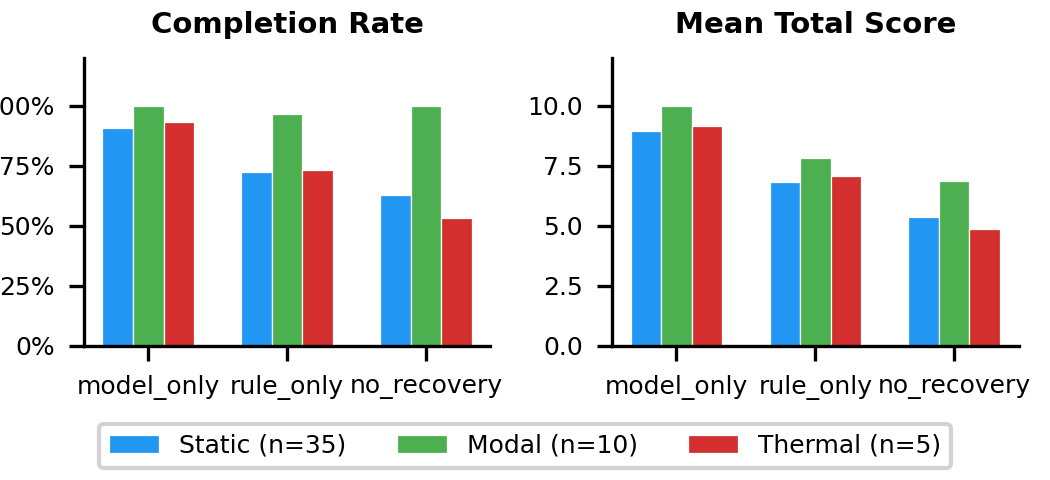}
  \caption{Task-type breakdown over static, modal, and thermal subsets. Model\_only keeps the strongest completion and score levels across all subsets, with the largest margin on thermal tasks.}
  \label{fig:type_breakdown}
\end{figure}

The task-type breakdown in Figure~\ref{fig:type_breakdown} shows that model\_only maintains the top completion and score levels on static, modal, and thermal subsets. The largest margin appears in thermal tasks, where no\_recovery drops to 0.5333 completion while model\_only stays at 0.9333.

\begin{figure}[t]
  \centering
  \includegraphics[width=\linewidth]{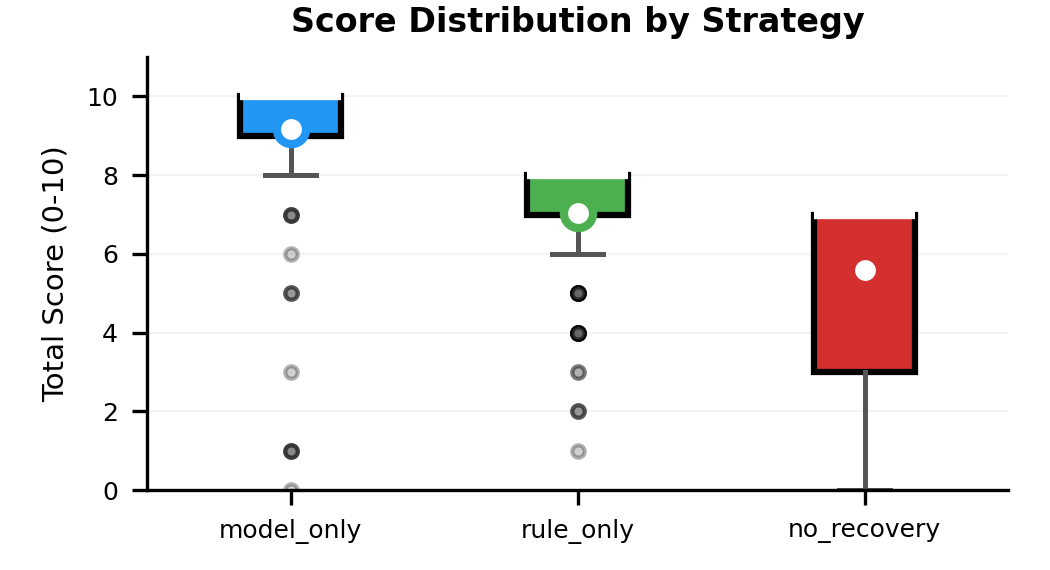}
  \caption{Interquartile score ranges and medians for each strategy. Model\_only concentrates near the top of the 0--10 scale, while no\_recovery shows lower and wider central tendency.}
  \label{fig:score_dist}
\end{figure}

Score distribution analysis in Figure~\ref{fig:score_dist} further indicates that model\_only is more robust: its quartiles are concentrated near the top of the 0--10 range, whereas no\_recovery shows a lower center and wider degradation.

Majority-case completion rates (across three repeats) are 0.94 for model\_only, 0.80 for rule\_only, and 0.72 for no\_recovery. Persistent hard-case IDs are concentrated in rule\_only and no\_recovery, suggesting deterministic edits alone are insufficient for long-tail solver and script-path failures.

\subsection{Failure-Case Analysis and Discussion}
Failure distribution is not uniform across simulation categories. At majority-case level, model\_only has 3 failed cases, all in static analysis (3 static, 0 thermal, 0 modal). Rule\_only has 10 failed cases (9 static, 1 thermal, 0 modal). no\_recovery has 14 failed cases (12 static, 2 thermal, 0 modal). This indicates that most residual failures are concentrated in static structural settings.

\begin{table}[t]
\centering
\caption{Failed-case distribution by strategy and simulation type (majority-case criterion).}
\label{tab:failure_dist}
\begin{tabular}{lcccc}
\toprule
Strategy & Total Failed & Static & Thermal & Modal \\
\midrule
model\_only  & 3  & 3  & 0 & 0 \\
rule\_only   & 10 & 9  & 1 & 0 \\
no\_recovery & 14 & 12 & 2 & 0 \\
\bottomrule
\end{tabular}
\end{table}

\begin{figure}[t]
  \centering
  \includegraphics[width=\linewidth]{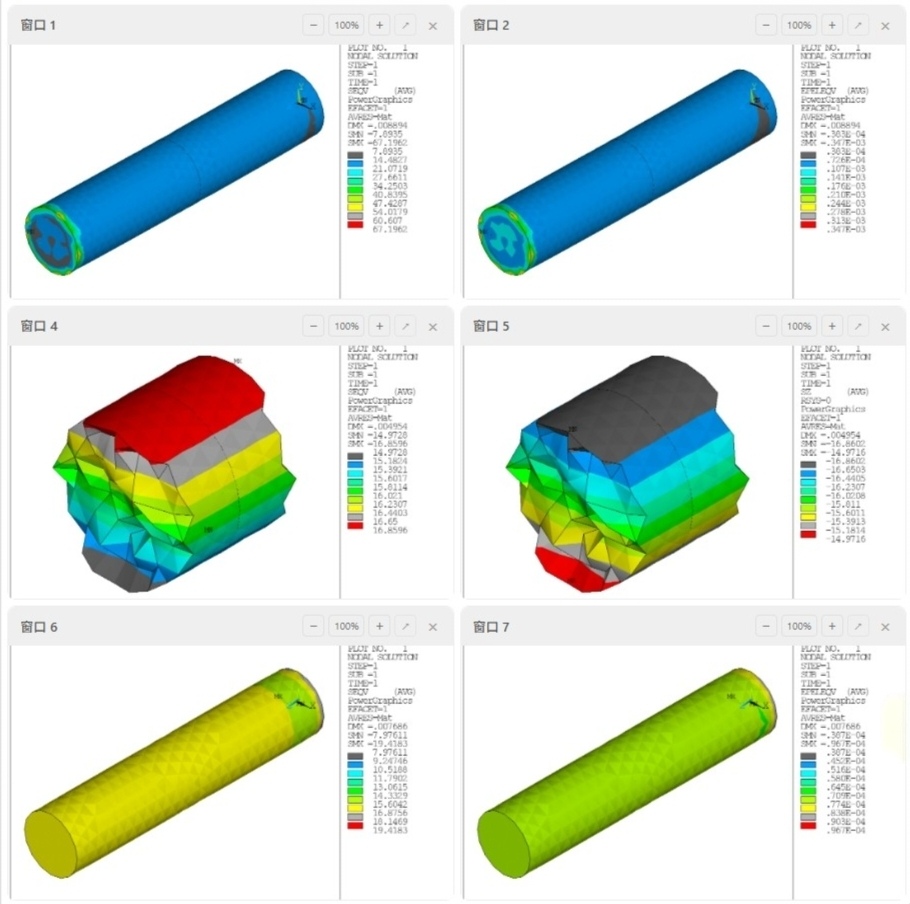}
  \caption{Mesh re-partition and re-meshing workflow. The visual pipeline indicates improved mesh quality after reprocessing, which is critical for reducing mesh-driven solver failures.}
  \label{fig:remesh_flow}
\end{figure}

For model\_only, the remaining failed static cases are concentrated in case IDs 8, 21, and 35. Combined with manual inspection, these failures are more correlated with thin-wall or mesh-sensitive geometric features than with nominal task complexity. In other words, some multi-part assemblies can still be solved successfully, while certain thin-wall geometries remain brittle due to mesh handling quality. Figure~\ref{fig:remesh_flow} illustrates the mesh re-partitioning and re-meshing pipeline proposed to address such geometry-driven failures.

\begin{figure}[t]
  \centering
  \includegraphics[width=0.95\linewidth]{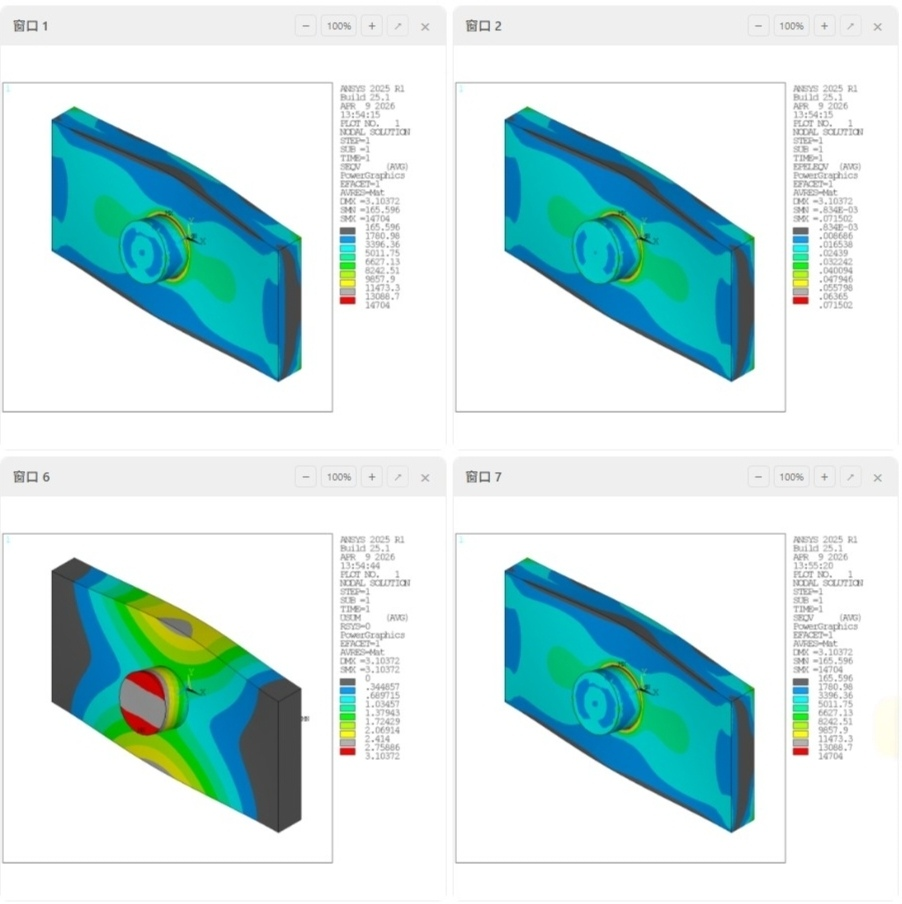}
  \caption{Static pin-joint case (\(100\times50\times10\) plate with a \(\phi20\) hole and \(\phi20\) pin under 2000N lateral load). This example illustrates that failure is tied to geometry-meshing sensitivity rather than simply the number of parts in the assembly.}
  \label{fig:pin_joint_case}
\end{figure}

The empirical pattern supports a simple interpretation: deterministic rules are useful but bounded by pre-specified templates, while model-driven repair adapts to broader failure signatures. Importantly, model\_only improves both completion and autonomy, indicating that gains are not achieved by shifting burden to human operators. For current CAX deployment, this places model\_only as the most practical default strategy.

\section{Limitations}
Several limitations of this study should be noted. First, the benchmark uses 50 structurally simple geometries (beams, plates, and cylinders with regular cross-sections). It does not include complex assemblies, nonlinear material models, large-deformation analysis, or multiphysics coupling. Results may not transfer directly to these more challenging settings. Second, the experiment is confined to a single solver backend (MAPDL) and a single external model (Claude Sonnet 4.6). Recovery behavior may differ with other solver interfaces or model versions. Third, the rule set used in rule\_only was derived from an internal engineering analysis of MAPDL failure modes rather than a systematic, exhaustive taxonomy; its effectiveness is bounded by the coverage of the failure patterns analyzed. Fourth, we report three repeated runs per strategy; larger repetition counts would yield tighter confidence intervals and more robust failure-mode statistics. Fifth, this study isolates the recovery policy as the experimental variable; other harness components shown in the architecture (Context Manager, Tool Pipeline, State Tracker) are part of the system design but are not individually ablated. Their contributions remain to be quantified in future work. Sixth, we do not directly compare CAX-Agent against FEABench, AutoFEA, or other recent FEA automation systems because these systems target different solver backends and task distributions; cross-system comparison on a common benchmark is an important direction for future work. Seventh, the recovery ladder shares conceptual similarity with the three-level hierarchy recently formalized in the Dual-State Action Pair framework \cite{ref26}. While this parallel supports the generality of layered recovery in agent harnesses, the present study does not provide a head-to-head comparison with Thompson's framework, and both that work and the harness-theory references \cite{ref05,ref06} are currently available only as arXiv preprints; their conclusions should be interpreted with appropriate caution pending peer-reviewed publication.

These limitations bound the scope of our conclusions but do not invalidate the core finding: under controlled conditions, model-driven recovery outperforms both no-recovery and rule-based repair on the metrics reported.

\section{Conclusion}
This paper presented CAX-Agent, a lightweight agent harness for MAPDL-based finite-element automation, and evaluated its recovery component through a controlled comparison of three strategies on 50 standardized structural benchmarks with repeated runs, blind human scoring, and inter-rater validation. Model\_only achieved the best results across all metrics: completion rate 0.9267, task score 3.59/4, total score 9.16/10, and zero-intervention rate 0.84, with large and statistically significant pairwise gains over rule\_only and no\_recovery. The zero-intervention gap between model\_only (0.84) and rule\_only (0.00) is particularly notable: deterministic rules, while improving completion over no recovery, never reached fully autonomous operation, underscoring the value of model-driven recovery within a harness architecture.

CAX-Agent demonstrates that a lightweight, domain-native harness---rather than a generic multi-agent framework---can transform MAPDL simulation from scattered, unreliable LLM calls into a standardized, traceable, and repeatable engineering workflow. The three-layer design, the recovery ladder, and the orchestrator-centric control model address the key failure modes that prevent reliable single-model deployment in mechanical simulation.

The benchmark scope is deliberately narrow: simple geometries, linear elasticity, single-physics. Validation on larger and more diverse benchmarks, across multiple solver backends and model versions, is needed before practical deployment. Extending recovery into the pre-processing stage---through adaptive meshing or geometry-aware partitioning---may address the mesh-sensitive failures that persist under script-level repair. From a harness-engineering perspective, the recovery-policy evaluation methodology used here---controlled ablation, repeated runs, multi-axis scoring with inter-rater validation---can be applied to other harness components and other simulation backends, offering a reproducible template for empirical harness evaluation as the field matures.


\end{document}